\documentclass{article}
\usepackage{ijcai21}

\usepackage{times}
\usepackage{soul}
\usepackage{url}
\usepackage[hidelinks]{hyperref}
\usepackage[utf8]{inputenc}
\usepackage[small]{caption}
\usepackage{graphicx}
\usepackage{amsmath}
\usepackage{amsthm}
\usepackage{amssymb}
\usepackage{multirow}
\usepackage{bigstrut}
\usepackage{booktabs}
\usepackage{algorithm}
\usepackage{algorithmic}
\usepackage{amsfonts}
\usepackage{subfigure}
\usepackage{color}

\title{Center Prediction Loss for Re-identification}
 
\author{
	Lu Yang$^1$
	\and
	Yunlong Wang$^1$\and
	Lingqiao Liu$^{3}$\and
	Peng Wang*$^{1}$\and
	Lu Chi$^{2}$\and
	Zehuan Yuan$^{2}$\and \\
	Changhu Wang$^{2}$\and
	Yanning Zhang$^{1}$
	\affiliations
	$^1$Northwestern Polytechnical University\and $^2$Bytedance AI Lab\and $^3$University of Adelaide
	\emails
	\{lu.yang, wangyunlong2019\}@mail.nwpu.edu.cn,
	lingqiao.liu@adelaide.edu.au,
	\{peng.wang, ynzhang\}@nwpu.edu.cn,
	\{chilu, yuanzehuan, wangchanghu\}@bytedance.com
}

\begin{document}

\maketitle

\begin{abstract}

The training loss function that enforces certain training sample distribution patterns plays a critical role in building a re-identification (ReID) system. Besides the basic requirement of discrimination, i.e., the features corresponding to different identities should not be mixed, additional intra-class distribution constraints, 
such as features from the same identities should be close to their centers, have been adopted to construct losses. Despite the advances of various new loss functions, it is still challenging to strike the balance between the need of reducing the intra-class variation and allowing certain distribution freedom. In this paper, we propose a new loss based on \emph{center predictivity}, that is, a sample must be positioned in a location of the feature space such that from it we can roughly predict the location of the center of same-class samples. The prediction error is then regarded as a loss called \textbf{C}enter \textbf{P}rediction \textbf{L}oss (CPL).
We show that, without introducing additional hyper-parameters, this new loss leads to a more flexible intra-class distribution constraint while ensuring the between-class samples are well-separated. Extensive experiments on various real-world ReID datasets show that the proposed loss can achieve superior performance and can also be complementary to existing losses.
\end{abstract}


\paragraph{}

    
 
\section{Introduction}

Re-identification (ReID), as an important application of deep metric learning, aims to retrieve the images with the same identity of queries. The common solution is to construct an embedding space such that samples with identical identities (IDs) are gathered while samples with different IDs are well separated. To enforce such a property, loss functions that measure the quality of an embedding distribution plays a critical role and underpins the success of a ReID system.

Generally speaking, the loss function for ReID systems usually involves two terms: between-class discriminative losses and intra-class losses. The former encourages the separation of samples with different IDs, e.g., triplet loss \cite{schroff2015facenet}, ID entropy loss, circle loss \cite{sun2020circle} etc. Intra-class losses are applied to samples within the same ID, their main purpose is to reduce the intra-class variation of samples. For example, center loss \cite{wen2016discriminative} enforces all the intra-class samples to be clustered around a learned class-specific center.
It try to shrink samples of the same class into one point in the feature space and may easily drop their similarity structure~\cite{wang2019ranked}.
The empirical study shows that applying intra-class losses like center-loss can improve the generalization performance of learned metric in applications like face recognition \cite{wen2016discriminative}. However, for applications with higher intra-class variations like Person-ReID, 
center loss can boost the clustering effect of features, but may reduce the ranking performance of ReID models~\cite{luo2019bag}.
This is because intra-class samples can exhibit significant intra-class variation and may not be characterized by a single center~\cite{sohoni2020no} or other existing simple distribution prior. 

\begin{figure}[tb]
\centering 
\includegraphics[scale=0.67]{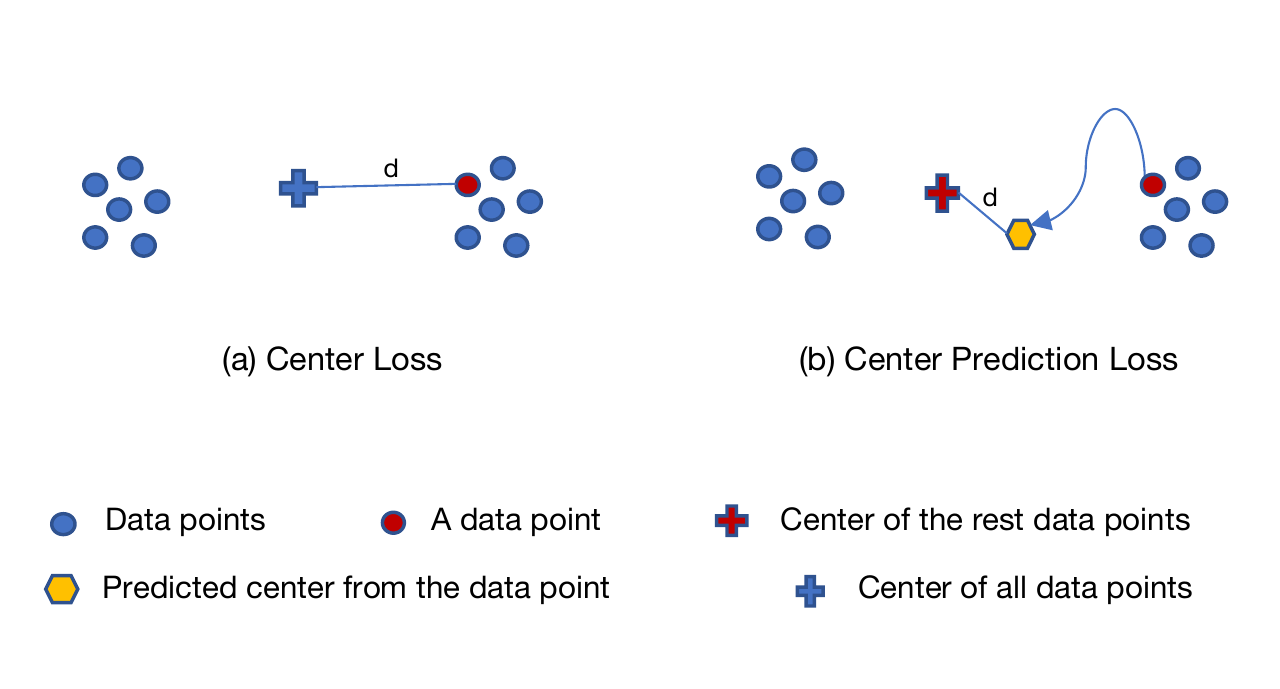} 

\caption{An example illustrates the difference between the center loss and the center prediction loss. Assuming that the intra-class data distribution is a Gaussian mixture with two Guassian components, the center loss and center prediction loss incur different loss values. (a) For the center loss, since each data point is far from the center, a high loss value will be incurred. (b) For the center prediction loss, a point (red) far from the center can be mapped to a point close to the center of the remaining points through the learned MLP. Therefore, a much smaller loss will be incurred and more flexible similarity structure can be preserved inside the class. \textbf{Best viewed in color.}}
\label{fig:CPL_vs_center_loss}
\end{figure}

To overcome the issue of the current intra-class losses, in this paper we propose a new loss called the center prediction loss to strike the balance between intra-class variation reduction and flexible intra-class distribution modeling. The main idea is to regularize the intra-class data distribution via a simple requirement: a sample must be positioned in a location such that from it we can roughly predict the location of the center of remaining samples. The prediction error is treated as a loss function called center prediction loss. We show that using the predictor in the CPL allows more freedom in choosing intra-class distribution families and the CPL can naturally preserve the between-class discrimination. 


By conducting extensive experiments on various person ReID and vehicle ReID datasets, we demonstrate the superior performance of the proposed loss and show that it can lead to consistent performance improvement to several existing ReID systems.

In summary, the contribution of this study is three-fold:
\begin{itemize}
\item 
We propose a new intra-class loss called \textbf{C}enter \textbf{P}rediction \textbf{L}oss (CPL). To the best of our knowledge, it is the first attempt to use the property of center predictivity as the loss function.

\item We show that the CPL allows more freedom for choosing the intra-class distribution family and can naturally preserving the discrimination between samples from different classes.

\item Extensive experiments on various ReID benchmarks show that the proposed loss can achieve superior performance and can also be complementary to existing losses. We also achieve new state-of-the-art performance on multiple ReID benchmarks.
\end{itemize}

\section{Related Work}
This section reviews some commonly used loss function for metric learning. 

\noindent\textbf{ID cross-entropy loss} treats each ID as an class, and use the cross-entropy loss to perform multi-class classification:
\begin{equation}
\mathcal{L}_{ce} =-\mathbb{E}_{\mathbf{x}} \left[ \log \frac{\exp \left(\mathbf{w}_{y}^{\top} \mathbf{x}+b_{y}\right)}{\sum_{j}^{C} \exp \left(\mathbf{w}_{j}^{\top} \mathbf{x}+b_j\right)} \right],
\end{equation}
where $\mathbb{E}$ is the expectation. $\mathbf{x}$ and $y$ are a randomly sampled embedding feature and its class label.  $\left[\mathbf{w}_{1}, \cdots, \mathbf{w}_{C}\right] \in \mathbb{R}^{d \times C}$is the weights of the classifier and $b$ is the bias term. In practice, we approximate the $\mathbb{E}$ with mini-batch in the SGD optimization.

\noindent\textbf{Center Loss}~\cite{wen2016discriminative} simultaneously learns a center for each class and penalizes the distances between the deep features and their corresponding class centers.
\begin{equation}
\mathcal{L}_{center} =\frac{1}{2} \mathbb{E}_{\mathbf{x}} \left[ \left\|\boldsymbol{x}-\boldsymbol{c}_{y}\right\|_{2}^{2}\right],
\end{equation}
where $\boldsymbol{c}_{y}$ is the learned center for class $y$. The definition of other symbols are same as above. 
Working with the ID cross-entropy loss, center loss can obtain excellent intra-class compactness and inter-class separability on the training set. 


\noindent\textbf{Triplet Loss}~\cite{schroff2015facenet} applies to a triplet of samples called anchor point ($\mathbf{x}_a$), positive point ($\mathbf{x}_p$) and negative point ($\mathbf{x}_n$). It aims to pull an anchor point closer to the positive point (same identity) than to the negative point (different identity) by a fixed
margin $m$:
\begin{equation}
\begin{split}
\mathcal{L}_{tri}= \mathbb{E}_{\{\mathbf{x}_a,\mathbf{x}_p, \mathbf{x}_n\}}\left[
D(\mathbf{x}_a,\mathbf{x}_p) - D(\mathbf{x}_a,\mathbf{x}_n) + m\right]_{+},
\end{split}
\label{reward}
\end{equation}
where $D(\cdot)$ denotes euclidean distance, $m$ is a fixed margin and $[\cdot]_+$ is the hinge function.

\noindent\textbf{Circle Loss}~\cite{sun2020circle} applies to a randomly sampled batch of samples. Given a single sample $\mathbf{x}$ in the feature space, assume that there are $K$ same-class samples which gives $K$ similarity scores and $L$ other-class sample which gives $L$ between-class similarity scores, then circle loss is defined as:
\begin{equation}
\resizebox{0.48\textwidth}{!}{
$\begin{aligned}
\mathcal{L}_{circle} 
&= \mathbb{E} \left[ \log \left(1+\sum_{j=1}^{L} \exp \left(\gamma\left(s_{n}^{j}+m\right)\right) \sum_{i=1}^{K} \exp \left(\gamma\left(-s_{p}^{i}\right)\right)\right) \right],
\end{aligned}
$}
\end{equation}
where $s_{n}^{j}$ is the intra-class similarity, $s_{p}^{i}$ is the between-class similarity. $m$ is the margin between $s_{n}^{j}$ and $s_{p}^{i}$ and $\gamma$ is a hyper-parameters.

\noindent\textbf{Lifted Structure Loss}~\cite{oh2016deep}
tries to pull one positive pair as close as possible and pushes all negative samples farther than a margin $m$.
\begin{equation}
\resizebox{0.49\textwidth}{!}{
$
\begin{aligned}
\mathcal{L}_{LS}= \mathbb{E} \left[ \sum_{y_{i j}=1}\left(D_{i j}+\log \left(\sum_{y_{i k}=0} \exp \left(m-D_{i k}\right)\right)+\log \left(\sum_{y_{j l}=0} \exp \left(m-D_{j l}\right)\right)\right)_{+} \right],
\end{aligned}
$
}
\end{equation}
where $D(\cdot)$ is the distance metric, $ij$ is positive pair, $ik$ and $jl$ are negative pairs.

\noindent\textbf{Ranked List Loss (RLL)}~\cite{wang2019ranked} proposes to build a set-based similarity structure by exploiting all instances in the gallery. Different from the above methods which aim to pull positive pairs as close as possible in the embedding space,
the RLL only needs to pull positive examples closer than a predefined threshold (boundary). In the form of pairwise constraint, the loss can be expressed as:
\begin{equation}
\mathcal{L}_{RLL}=\mathbb{E} \left[ \left(1-y_{i j}\right)\left[\alpha-d_{i j}\right]_{+}
+ y_{i j}\left[d_{i j}-(\alpha-m)\right]_{+} \right],
\end{equation}
where $d_{ij}$ is euclidean distance between $i$th sample and $j$th sample. $y_{i j}=1$ if $y_{i}=y_{j}$, and $y_{i j}=0$ otherwise. RLL aims to push its negative point farther than a boundary $\alpha$ and pull its positive one closer than another boundary $\alpha-m$. Thus $m$ is the margin between two boundaries.

\noindent\textbf{N-pair-mc Loss} \cite{sohn2016improved} uses the structural information between the data to learn more discriminative features. Triplet loss pulls one positive point towards anchor while pushing a
negative one simultaneously. In the process of each parameter update, N-pair-mc loss also considers the relationship between the query sample and negative samples of different classes, prompting the query to maintain a distance from all other classes, which can speed up the convergence of the model.
\noindent\textbf{Limitations of the existing losses:} By analyzing the above loss functions, we identify the following limitations of the existing methods:
First, all of the above loss functions try to reduce the distance between samples of the same class. Most losses encourage the same-class samples to be shrunken into one point. However, this may hurt preserving the intrinsic intra-class sample diversity which has been shown to be beneficial for learning transferable features \cite{wang2019ranked}. 
Second, some loss functions, such as RLL, use a fixed intra-class margin for all classes to regularize the intra-class sample distribution. However, a fixed margin can hardly be optimal for all classes.

In contrast, our proposed CPL allows more flexibility in intra-class sample distribution and does not necessarily require same-class samples should have very small distances. Also, it does not involve any pre-set hyper-parameters and can be adaptive to each class.




\begin{figure}[tb]
\begin{center}
    \includegraphics[width=0.45\textwidth]{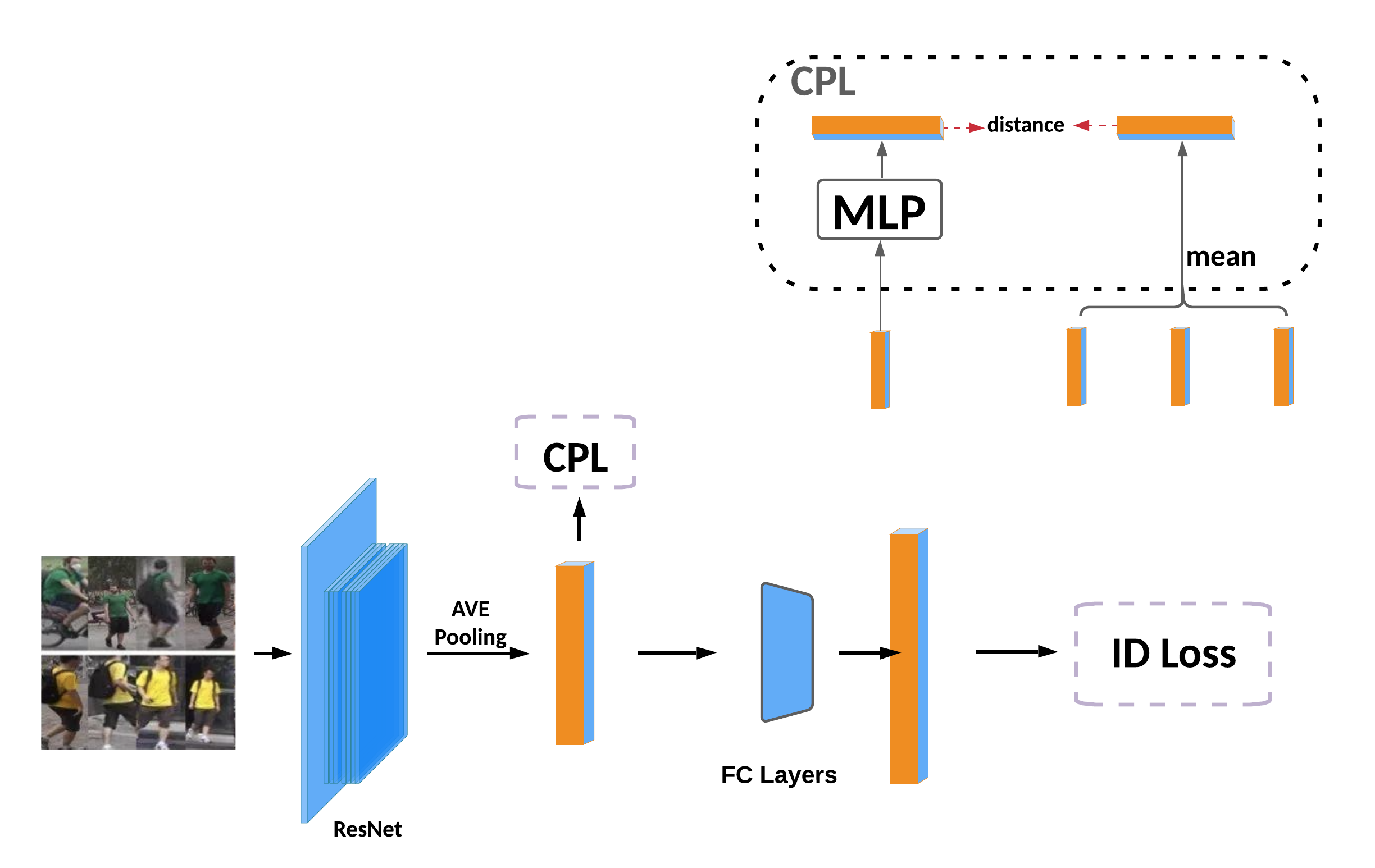}
\end{center}
\caption{A default way of applying the proposed CPL. The CPL is an intra-class loss and it needs to be used with another between-class discriminative loss. In this approach, we apply a fully connection layer as an identity classifier and use the ID cross-entropy loss.}
\label{fig:cifar_architecture}
\end{figure}

\section{Approach}

In this section, we elaborate on our approach. We first introduce the detail of the proposed \textbf{C}enter \textbf{P}rediction \textbf{L}oss (CPL). Then we explain the properties of the CPL with examples. 
Finally, we provide implementation details of the CPL.

\begin{figure}[th]
\centering 
\includegraphics[scale=0.4]{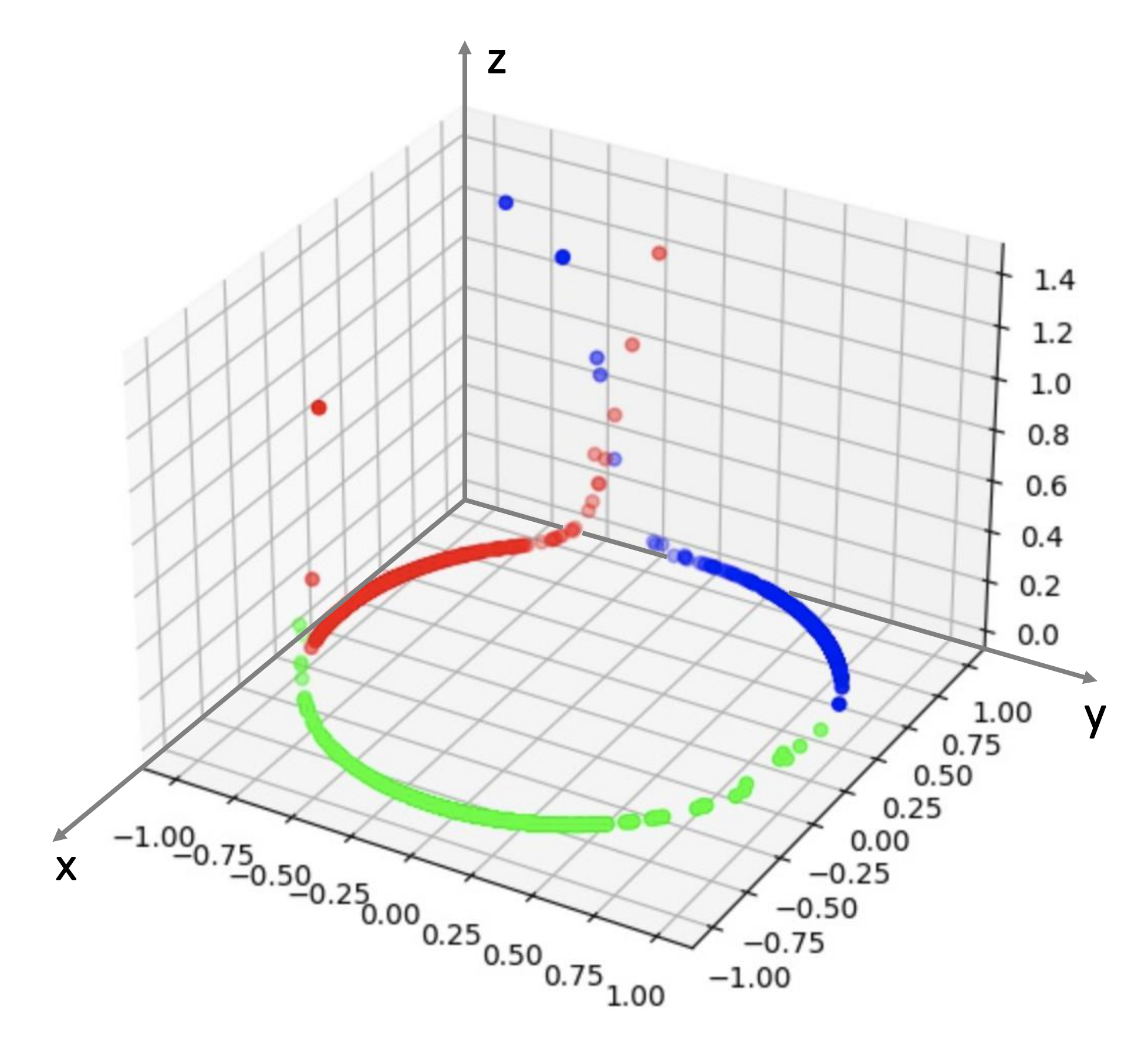} 
\caption{The learned $2$D embeddings ($L$-2 normalized) and their corresponding CPL (z-axis) in CIFAR-10 experiment. Different colors denote different classes. Best viewed in color. As can be seen, samples located at the class boundary tend to incur higher CPL.}
\label{fig:toy_visualization}
\end{figure}

\begin{figure*}[ht!]
\begin{center}
    \includegraphics[width=0.8\textwidth]{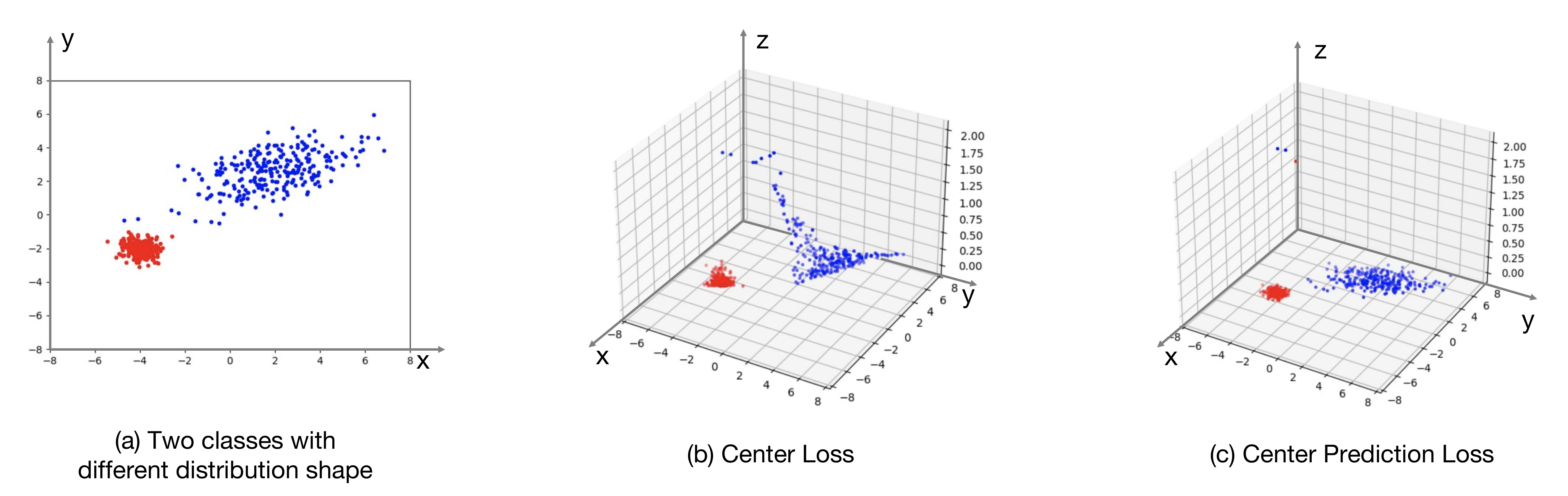}
\end{center}
\caption{The center loss and the center prediction loss (CPL) on two classes with different Gaussian distributions. The class (red samples) on the left is a compact circular distribution, and the class (blue samples) on the right is an elliptic distribution with large co-variance. \textbf{Best viewed
in color}.}
\label{fig:lossshape}
\end{figure*}
\subsection{Center Prediction Loss}

The proposed CPL is calculated from a set of randomly sampled same-class samples. After extracting features from a feature extractor $\phi(\cdot)$, we obtain a set of feature vectors $\{\mathbf{x}_1,\mathbf{x}_2, \cdots,\mathbf{x}_k\}$ from those intra-class samples. The CPL is defined as:
\begin{align}
\label{Fun:eq_CPL}
    \mathcal{L}_{CPL} = \min_{\theta}\mathbb{E}_{\{\mathbf{x}_1,\cdots,\mathbf{x}_k\}}\left[\sum_{i=1}^{k} \|f(\mathbf{x}_i;\theta) - \mathbf{c}_i\|_2^2\right],
\end{align} 
where $k$ is the number of intra-class samples, $f(\mathbf{x}_i;\theta)$ is the center predictor, which consists of a multi-layer perceptron (MLP) parameterized by $\theta$. $\mathbf{c}_i = \frac{1}{k-1} \sum_{j\neq i} \mathbf{x}_j$ \footnote{There might be other choices for the prediction target $\mathbf{c}_i$, e.g., the center of all same-class samples or even a randomly sampled same-class data point. We empirically find that using the center of the remaining samples leads to slightly better performance. Investigation of other target choices is provided in our ablation study.}. $\mathbb{E}$ denotes the expectation. In other words, as shown in Figure \ref{fig:cifar_architecture}, CPL is defined as the optimal predictive error from a sample to the center of the rest of same-class samples. At the first glance, CPL is very similar to the center loss, which directly requires $\mathbf{x}_i$ to be close to a learned center. However, we will show that the introduction of the center predictor (MLP) significantly changes the properties of the loss. 

\subsection{Properties of the CPL}
The difference between CPL and traditional single-center-based intra-class loss can be well explained with the example shown in Figure \ref{fig:CPL_vs_center_loss}. As seen, for a Guassian-mixture-alike distribution shown in Figure \ref{fig:CPL_vs_center_loss}, the center loss will incur heavy penalization since the samples are not gathered around a center. In contrast, CPL will not incur a heavy loss. This is because if a sample is located at the right (or the left) side Gaussian component, it should be aware that the center of the rest same-class samples should be located in the middle. Thus, the center predictor will project a sample to the middle of the two Gaussians and lead to a small CPL. From this example, we can see that the CPL allows greater flexibility for the distribution formation. 

One potential concern of the CPL is that if it allows too much freedom for the intra-class distribution it may hurt the discrimination of between-class samples. In fact, the center prediction mechanism can automatically avoid positioning samples at the class boundary. This is because, for boundary samples,  a slight perturbation of samples will lead to a significant change of the prediction target which may jump from the center of one class sample to that of another class sample. Therefore, the center predictor will struggle to make the right prediction and in effect, it tends to incur high loss values. To validate this claim, we conduct an experiment on a subset of CIFAR-10. We set the output number of the last hidden layer to $2$ in ResNet20, which allows us to plot the features on $2$-D surface for visualization. We learn 2-dimension embeddings of samples with both an ID entropy loss and the CPL. 
The results are shown in Figure~\ref{fig:toy_visualization} in which different colors are used to denote samples from different classes and values of CPL are shown on the $z$-axis. As can be seen, the points at the class-boundary lead to significantly higher loss values.   

We further use a simple example to illustrate how CPL works in practice. In Figure~\ref{fig:lossshape} (a), there are two Gaussian distributions with different covariance matrices, representing two classes. Then we train a center predictor for CPL, and calculate the prediction error, $e_i = \|f(\mathbf{x}_i;\theta) - \mathbf{c}_i\|_2^2$ for every data point (Note that $\mathcal{L}_{CPL} = \mathbb{E}\left[\sum e_i\right]$). Then we use the $z$-axis of each point to indicate $e_i$. To make a comparison with the center loss, we use the same visualization method but calculating $e_i = \|\mathbf{x}_i - \mathbf{c}\|_2^2$, where $\mathbf{c}$ is the respective class mean vector. The visualization results are shown in Figure~\ref{fig:lossshape} (b) and (c). From the visualization, we can see that the center loss penalizes the blue-class points much heavier than the red-class points since the former has a relatively large variance. In contrast, the CPL does not over-penalize most blue-class points. The only high loss values are incurred at the points located at the class boundary. Therefore, minimizing the CPL will not strongly force blue points to move closer to the center but push points away from the class boundary.

\subsection{Implementation of CPL}
To straightforwardly implement the CPL, we need first update the center predictor to find the best fit to the target and then back-propagate through the loss to the feature extractor of $\mathbf{x}_i$. However, this is computationally intractable. In practice, we \textit{jointly optimize} the parameters inside the center predictor and the other network parameters. To avoid co-adaptation of $\mathbf{x}$ and $f(\cdot)$, e.g., the network may push $\mathbf{x}$ to a very small scale for minimizing the CPL, we \textit{does not allow gradient update through $\mathbf{c}_i$.} In other words, we fix $\mathbf{c}_i$ as constant once it is calculated in the forward pass. Another important implementation detail is that we apply a batch normalization to $\mathbf{x}_i$ when calculating $\mathbf{c}_i$, namely, $\mathbf{c}_i = \frac{1}{K-1} \sum_{j\neq i} BN(\mathbf{x}_j)$. 
Since intra-class loss cannot be used for metric learning alone, we by default to apply it with the ID cross-entropy loss, that is, a classifier is attached to the learned embedding to categorize a sample to its given ID. Figure \ref{fig:cifar_architecture} shows the structure of the system, where we use a classifier for ID classification. 

Note that the proposed CPL can be simply plugged into any deep metric learning system as an additional loss function. We present the pseudo-code of the CPL in Algorithm \ref{alg}. 
\begin{center}
    \vskip -0.1in
        \begin{algorithm}[H]
            \caption{Applying Center Prediction Loss for ReID systems}
            \label{alg}
        \begin{algorithmic}[1]
            \STATE {\bfseries Input:} $\mathcal{D}$
            \STATE Randomly initialize feature extractor $\phi(\cdot;\lambda)$ and center predictor (MLP) $f(\cdot;\theta)$
            
            \FOR{$t=0$ {\bfseries to} $T$}
            \STATE Sample $P$ IDs with $K$ images per ID from $\mathcal{D}$. Extracting their features as $\{\{ \mathbf{x}_i^c \}_{i=1}^K\}\}_{c=1}^P$  
            \STATE Compute $\mathbf{c}_i^c = \frac{1}{K-1} \sum_{j=1, j\neq i}^{K}  BN(\mathbf{x}_j^c)$
            \STATE Compute loss $\mathcal{L}_{CPL} = \sum_c \frac{1}{K} \sum_i \|f(\mathbf{x}_i^c;\theta) - \mathbf{c}_i\|_2^2$ 
            \STATE Calculate ID cross-entropy loss $\mathcal{L}_{ce}$ for samples
            \STATE Update both $\theta$ and $\lambda$ with by back-propagate $\mathcal{L}_{CPL} + \lambda \mathcal{L}_{ce}$ via SGD.
            \ENDFOR
 
        \end{algorithmic}
        \end{algorithm}
\end{center}

\begin{table*}[tb]
		\centering
		\footnotesize
	\scalebox{0.98}{
    \begin{tabular}{crrrrrrrrrrr}
    	\hline
    	\multicolumn{1}{c}{\multirow{3}[4]{*}{Method}} & \multicolumn{4}{c}{CUHK03} & \multicolumn{2}{c}{\multirow{2}[2]{*}{Market1501}} & \multicolumn{2}{c}{\multirow{2}[2]{*}{Duke}} & \multicolumn{2}{c}{\multirow{2}[2]{*}{MSMT17}} \bigstrut[t]\\
    	\multicolumn{1}{c}{} & \multicolumn{2}{c}{Labeled} & \multicolumn{2}{c}{Detected} & \multicolumn{2}{c}{} & \multicolumn{2}{c}{} \bigstrut[b]\\
    	\cline{2-11} \multicolumn{1}{c}{} & \multicolumn{1}{c}{Rank-1} & \multicolumn{1}{c}{mAP} & \multicolumn{1}{c}{Rank-1} & \multicolumn{1}{c}{mAP} & \multicolumn{1}{c}{Rank-1} & \multicolumn{1}{c}{mAP} & \multicolumn{1}{c}{Rank-1} & \multicolumn{1}{c}{mAP} & \multicolumn{1}{c}{Rank-1} & \multicolumn{1}{c}{mAP} \bigstrut\\
    	\hline
    	 
    	\multicolumn{1}{l}{IDE~\cite{sun2017beyond}} & \multicolumn{1}{c}{43.8 } & \multicolumn{1}{c}{38.9 } & \multicolumn{1}{c}{-} & \multicolumn{1}{c}{-} & \multicolumn{1}{c}{85.3 } & \multicolumn{1}{c}{68.5 } & \multicolumn{1}{c}{73.2 } & \multicolumn{1}{c}{52.8 } & \multicolumn{1}{c}{-} & \multicolumn{1}{c}{- } \bigstrut[t]\\ 
    	 
    	\multicolumn{1}{l}{MGCAM~\cite{song2018mask}} & \multicolumn{1}{c}{50.1 } & \multicolumn{1}{c}{50.2 } & \multicolumn{1}{c}{46.7 } & \multicolumn{1}{c}{46.9 } & \multicolumn{1}{c}{83.8 } & \multicolumn{1}{c}{74.3 } & \multicolumn{1}{c}{-} & \multicolumn{1}{c}{-}& \multicolumn{1}{c}{-} & \multicolumn{1}{c}{- } \bigstrut[t]\\
    	\multicolumn{1}{l}{AACN~\cite{xu2018attention}} & \multicolumn{1}{c}{-} & \multicolumn{1}{c}{-} & \multicolumn{1}{c}{-} & \multicolumn{1}{c}{-} & \multicolumn{1}{c}{85.9 } & \multicolumn{1}{c}{66.9 } & \multicolumn{1}{c}{76.8 } & \multicolumn{1}{c}{59.3 } & \multicolumn{1}{c}{-} & \multicolumn{1}{c}{- }\bigstrut[t]\\
    	\multicolumn{1}{l}{SPReID~\cite{kalayeh2018human}} & \multicolumn{1}{c}{-} & \multicolumn{1}{c}{-} & \multicolumn{1}{c}{-} & \multicolumn{1}{c}{-} & \multicolumn{1}{c}{92.5 } & \multicolumn{1}{c}{81.3 } & \multicolumn{1}{c}{84.4 } & \multicolumn{1}{c}{71.0 }  & \multicolumn{1}{c}{-} & \multicolumn{1}{c}{- }\bigstrut[t]\\
    	\multicolumn{1}{l}{HA-CNN~\cite{li2018harmonious}} & \multicolumn{1}{c}{44.4 } & \multicolumn{1}{c}{41.0 } & \multicolumn{1}{c}{41.7 } & \multicolumn{1}{c}{38.6 } & \multicolumn{1}{c}{91.2 } & \multicolumn{1}{c}{75.7 } & \multicolumn{1}{c}{80.5 } & \multicolumn{1}{c}{63.8 } & \multicolumn{1}{c}{-} & \multicolumn{1}{c}{- }\bigstrut[t]\\
    	\multicolumn{1}{l}{DuATM~\cite{si2018dual}} & \multicolumn{1}{c}{-} & \multicolumn{1}{c}{-} & \multicolumn{1}{c}{-} & \multicolumn{1}{c}{-} & \multicolumn{1}{c}{91.4 } & \multicolumn{1}{c}{76.6 } & \multicolumn{1}{c}{81.8 } & \multicolumn{1}{c}{64.6 }  & \multicolumn{1}{c}{-} & \multicolumn{1}{c}{- }\bigstrut[t]\\
    	\multicolumn{1}{l}{Mancs~\cite{wang2018mancs}} & \multicolumn{1}{c}{69.0 } & \multicolumn{1}{c}{63.9 } & \multicolumn{1}{c}{65.5 } & \multicolumn{1}{c}{60.5 } & \multicolumn{1}{c}{93.1 } & \multicolumn{1}{c}{82.3 } & \multicolumn{1}{c}{84.9 } & \multicolumn{1}{c}{71.8 } & \multicolumn{1}{c}{-} & \multicolumn{1}{c}{- }\bigstrut[t]\\
    	\multicolumn{1}{l}{HPM~\cite{fu2019horizontal}} & \multicolumn{1}{c}{63.9 } & \multicolumn{1}{c}{57.5 } & \multicolumn{1}{c}{-} & \multicolumn{1}{c}{-} & \multicolumn{1}{c}{94.2 } & \multicolumn{1}{c}{82.7 } & \multicolumn{1}{c}{86.6 } & \multicolumn{1}{c}{74.3 }& \multicolumn{1}{c}{-} & \multicolumn{1}{c}{- } \bigstrut[t]\\
    	\multicolumn{1}{l}{DSA-reID~\cite{zhang2019densely}} & \multicolumn{1}{c}{78.9 } & \multicolumn{1}{c}{75.2 } & \multicolumn{1}{c}{78.2} & \multicolumn{1}{c}{73.1 } & \multicolumn{1}{c}{\textbf{95.7 }} & \multicolumn{1}{c}{87.6 } & \multicolumn{1}{c}{86.2 } & \multicolumn{1}{c}{74.3 }& \multicolumn{1}{c}{-} & \multicolumn{1}{c}{- } \bigstrut[t]\\
    	\multicolumn{1}{l}{IANet~\cite{zhang2019densely}} & \multicolumn{1}{c}{- } & \multicolumn{1}{c}{- } & \multicolumn{1}{c}{- } & \multicolumn{1}{c}{- } & \multicolumn{1}{c}{94.4 } & \multicolumn{1}{c}{83.1 } &
    	\multicolumn{1}{c}{- } & \multicolumn{1}{c}{- } &
    	\multicolumn{1}{c}{75.5 } & \multicolumn{1}{c}{46.8} \bigstrut[t]\\
    	\multicolumn{1}{l}{JDGL~\cite{fu2019horizontal}} & \multicolumn{1}{c}{- } & \multicolumn{1}{c}{- } & \multicolumn{1}{c}{-} & \multicolumn{1}{c}{-} & \multicolumn{1}{c}{94.8 } & \multicolumn{1}{c}{86.0 } &
    	\multicolumn{1}{c}{-} & \multicolumn{1}{c}{- } &
    	\multicolumn{1}{c}{77.2 } & \multicolumn{1}{c}{52.3 } \bigstrut[t]\\
    	\multicolumn{1}{l}{OSNet~\cite{fu2019horizontal}} & \multicolumn{1}{c}{- } & \multicolumn{1}{c}{- } & \multicolumn{1}{c}{72.3} & \multicolumn{1}{c}{67.8} & \multicolumn{1}{c}{94.8 } & \multicolumn{1}{c}{84.9 } & \multicolumn{1}{c}{- } & \multicolumn{1}{c}{- } & \multicolumn{1}{c}{78.7 } & \multicolumn{1}{c}{52.9 } \bigstrut[t]\\
        \multicolumn{1}{l}{HOReid~\cite{wang2020high}} & \multicolumn{1}{c}{-} & \multicolumn{1}{c}{-} & \multicolumn{1}{c}{-} & \multicolumn{1}{c}{-} & \multicolumn{1}{c}{94.2 } & \multicolumn{1}{c}{84.9} & \multicolumn{1}{c}{86.9} & \multicolumn{1}{c}{75.6} & \multicolumn{1}{c}{-} & \multicolumn{1}{c}{- } \bigstrut[t]\\
        
        \multicolumn{1}{l}{ISP~\cite{zhu2020identity}} & \multicolumn{1}{c}{76.5} & \multicolumn{1}{c}{74.1} & \multicolumn{1}{c}{75.2} & \multicolumn{1}{c}{71.4} & \multicolumn{1}{c}{95.3 } & \multicolumn{1}{c}{\emph{88.6}} & \multicolumn{1}{c}{\textbf{89.6}} & \multicolumn{1}{c}{\textbf{80.0}} & \multicolumn{1}{c}{-} & \multicolumn{1}{c}{- } \bigstrut[t]\\
    	\hline
    	\multicolumn{1}{l}{AlignedReID$^{\dag}$~\cite{luo2019alignedreid}} & \multicolumn{1}{c}{53.8} & \multicolumn{1}{c}{49.0} & \multicolumn{1}{c}{50.5} & \multicolumn{1}{c}{45.6} & \multicolumn{1}{c}{84.9} & \multicolumn{1}{c}{68.0} & \multicolumn{1}{c}{76.5} & \multicolumn{1}{c}{59.6}& \multicolumn{1}{c}{58.8} & \multicolumn{1}{c}{33.7}  \bigstrut[t]\\
    	\multicolumn{1}{l}{AlignedReID + CPL} & \multicolumn{1}{c}{55.6} & \multicolumn{1}{c}{50.1} & \multicolumn{1}{c}{52.4} & \multicolumn{1}{c}{47.0} & \multicolumn{1}{c}{85.2} & \multicolumn{1}{c}{69.5} & \multicolumn{1}{c}{79.0} & \multicolumn{1}{c}{61.3}& \multicolumn{1}{c}{62.7} & \multicolumn{1}{c}{36.5}  \bigstrut[b]\\
    	\hline
    	\multicolumn{1}{l}{BoT$^{\dag}$~\cite{luo2019bag}} & \multicolumn{1}{c}{68.9} & \multicolumn{1}{c}{66.8} & \multicolumn{1}{c}{65.2} & \multicolumn{1}{c}{62.9} & \multicolumn{1}{c}{93.7} & \multicolumn{1}{c}{84.9} & \multicolumn{1}{c}{85.2} & \multicolumn{1}{c}{75.1} & \multicolumn{1}{c}{73.8} & \multicolumn{1}{c}{50.1}  \bigstrut[t]\\
    	\multicolumn{1}{l}{BoT + CPL} &  \multicolumn{1}{c}{71.9} & \multicolumn{1}{c}{69.6} & \multicolumn{1}{c}{68.3} & \multicolumn{1}{c}{65.9} & \multicolumn{1}{c}{94.7} & \multicolumn{1}{c}{87.0} & \multicolumn{1}{c}{87.3} & \multicolumn{1}{c}{76.4} & \multicolumn{1}{c}{75.0} & \multicolumn{1}{c}{51.8} \bigstrut[b]\\
    	\hline
    	\multicolumn{1}{l}{RGA$^{\dag}$~\cite{zhang2020relation}} & \multicolumn{1}{c}{\emph{81.1}} & \multicolumn{1}{c}{\emph{77.2}} & \multicolumn{1}{c}{\emph{78.9}} & \multicolumn{1}{c}{\emph{75.0}} & \multicolumn{1}{c}{95.2} & \multicolumn{1}{c}{88.4} & \multicolumn{1}{c}{88.7} & \multicolumn{1}{c}{78.4} & \multicolumn{1}{c}{\emph{79.9}} & \multicolumn{1}{c}{\emph{56.7}} \bigstrut[t]\\
    	\multicolumn{1}{l}{RGA + CPL} & \multicolumn{1}{c}{\textbf{82.5}} & \multicolumn{1}{c}{\textbf{78.3}} & \multicolumn{1}{c}{\textbf{80.9}} & \multicolumn{1}{c}{\textbf{76.8}} & \multicolumn{1}{c}{\emph{95.5}} & \multicolumn{1}{c}{\textbf{88.8}} & \multicolumn{1}{c}{\emph{89.0}} & \multicolumn{1}{c}{\emph{78.8}} & \multicolumn{1}{c}{\textbf{81.4}} & \multicolumn{1}{c}{\textbf{57.8}} \bigstrut[b]\\

    	\hline
    \end{tabular}
    }

	\caption{Performance (\%) comparisons with the state-of-the-arts on CUHK03, Market1501, Duke and MSMT17. ``\dag'' means by our implementation. \textbf{Bold} and \emph{Italic} fonts represent the best and second best performance respectively.}
	\label{tab:SOTA_person}
\end{table*}

\begin{table}[tb]
	\begin{center}
		\scalebox{0.73}{
			\begin{tabular}{c|cc|cc}
				\hline
				\multicolumn{1}{c|}{\multirow{2}{*}{Method}} & \multicolumn{2}{c|}{VeRi-776} &  \multicolumn{2}{c}{VehicleID (Large)} \\ \cline{2-5} 
				\multicolumn{1}{c|}{} & \multicolumn{1}{c}{Rank-1} & \multicolumn{1}{c}{mAP}  & \multicolumn{1}{|c}{Rank-1} & \multicolumn{1}{c}{mAP}  \\
				\hline \hline
				AAVER$^*$ [Khorramshahi et al., 2019] & $89.0$  &  $61.2$ &$63.5$ & $-$\\
				PRND$^{*\ddag}$ [He et al., 2019] & $94.3$ & $74.3$ & $74.2$ & $-$\\
				\hline
				GS-TRE(ResNet50) [Bai et al., 2018] & $-$ & $-$ & $-$ & $78.8$\\
				FDA-Net [Lou et al., 2019b] & $84.3$ & $55.5$ & $55.5$ & $61.8$\\
				EALN [Lou et al., 2019a] & $84.4$ & $57.4$ & $69.3$ & $71.0$\\
				Mob.VFL [Alfasly et al., 2019] & $87.2$ & $58.1$ & $67.4$ & $-$\\
				QD-DLP [Zhu et al., 2019] & $88.5$ & $61.8$ & $64.1$ & $68.4$\\
				DMML [Chen et al., 2019] & $91.2$ & $70.1$ & $67.7^{\dag}$ & $72.4^{\dag}$\\
				MRL~\cite{lin2019multi}$^{\ddag}$& $94.3$ & $78.5$ & $\mathbf{78.4}$ & $\emph{81.2}$\\
				SAVER$^{\ddag}$~\cite{ECCV2020Devil}& $\mathbf{96.4}$ & $\emph{79.6}$ & $75.3$ & $-$\\
				\hline
				BoT$^{\dag}$~\cite{luo2019bag} & $94.7$ & $79.3$ & $75.5$ & $81.0$\\
				BoT + CPL & $\emph{96.0}$ & $\mathbf{80.6}$ & $\emph{76.7}$ & $\mathbf{82.3}$\\
				\hline
			\end{tabular}
	  	}
	\end{center}
	\caption{Comparison with state of the art methods on VeRi-776 and VehicleID (in \%). ``*'' indicates models trained with extra annotations such as keypoints or viewpoints. ``\dag'' means by our implementation. The input image size is $224\times 224$ by default except the methods with ``\ddag''. \textbf{Bold} and \emph{Italic} fonts represent the best and second best performance respectively. }
	\label{tab:SOTA_vehicle}
\vspace{-1em}
\end{table}

\begin{table}[tb]
\begin{center}
\scalebox{0.85}{
\begin{tabular}{c|cc}
\hline
\multicolumn{1}{c|}{Method} & Rank-1 & Rank-5 \\ \hline \hline
OIFE(Single Branch)* & $24.6$ & $51.0$ \\
Siamese-Visual* & $30.5$ & $57.3$ \\
MSVF~\cite{kanaci2018vric}& $46.6$ & $65.6$ \\
CRAD~\cite{li2020cross}& $50.1$ & $68.2$ \\
BW~\cite{kumar2020strong}& $69.1$ & $90.5$ \\
\hline
BoT$^{\dag}$~\cite{luo2019bag} & $74.3$ & $90.3$ \\ 
BoT + CPL    & $\mathbf{76.6}$ & $\mathbf{91.8}$ \\
\hline
\end{tabular}
}
 \caption{Comparison with state-of-the-art methods on VRIC (in \%). \textbf{Bold} fonts represent the best performance. ``\dag'' means by our implementation. ``*'' refers to results represented in MSVF.}
 \label{tab:SOTA_VRIC}
\end{center}
\vspace{-1mm}
\end{table}

\section{Experiments}

\subsection{Datasets}
We conduct extensive experiments on four public person ReID datasets, $i.e.$, Market1501~\cite{iccv15zheng}, Duke~\cite{iccv17duke}, CUHK03~\cite{cvpr14cuhk} and MSMT17~\cite{cvpr18msmt}. We also verify the effectiveness of our loss on three largescale vehicle ReID datasets, $i.e.$, VeRi-776~\cite{liu2016veri776}, VehicleID~\cite{liu2016vehicleid} and VRIC~\cite{kanaci2018vric}. The Cumulative Match Curve (CMC) at Rank-1 and the mean Average Precision (mAP) are used as the evaluation criteria.

\subsection{Implementation Details}
Unless specified otherwise, we use BOT~\cite{luo2019bag} as the baseline by default. We didn't utilize label smooth, test with flip and re-ranking. All the experiments use ResNet50 as the backbone. The center predictor is a two-layer MLP, whose hidden layer is $512$-dimension. The person images were resized to $256 \times 128$ and vehicle images were resized to $224 \times 224$. The batch size is $64$ ($4$ images/ID and $16$ IDs). The model is trained $120$ epochs and the learning rate is initialized to $3.5\times 10^{-4}$ and divided by $10$ at the $40$th epoch and $70$th epoch.
The detailed implementation of other methods follows the settings in their respective papers.

\subsection{CPL with SOTA Methods}
In this section we evaluate the performance of our proposed loss by incorporating it into several state-of-the-art approaches.
For person ReID benchmarks, we apply CPL on \cite{luo2019alignedreid,luo2019bag,zhang2020relation} that already use ID cross-entropy loss and hard triplet loss. The results are shown in Table~\ref{tab:SOTA_person}. It is clear that our CPL can further improve the performance of several SOTA methods on four public person ReID datasets.

To further validate the effectiveness of our loss, we also conduct experiments on three vehicle datasets.
Although BoT~\cite{luo2019bag} is originally designed for person ReID, applying it to vehicle ReID is straightforward given their similar pipelines. We resize the image size from $256 \times 128$ to $224 \times 224$ in vehicle ReID. As shown in Table~\ref{tab:SOTA_vehicle} and Table~\ref{tab:SOTA_VRIC}, our method yields a consistent performance gain over these vehicle datasets in different degrees, ranging from $1.2\%$ to $2.3\%$ on Rank-1.

\begin{table*}[tb]
\begin{center}
	\scalebox{0.9}{
    \begin{tabular}{crrrrrrrrrrr}
    	\hline
    	\multicolumn{1}{c}{\multirow{3}[4]{*}{Method}} & \multicolumn{4}{c}{CUHK03} & \multicolumn{2}{c}{\multirow{2}[2]{*}{Market1501}} & \multicolumn{2}{c}{\multirow{2}[2]{*}{Duke}} & \multicolumn{2}{c}{\multirow{2}[2]{*}{MSMT17}} \bigstrut[t] \\
    	\multicolumn{1}{c}{} & \multicolumn{2}{c}{Labeled} & \multicolumn{2}{c}{Detected} & \multicolumn{2}{c}{} & \multicolumn{2}{c}{} \bigstrut[b] \\
    	\cline{2-11} \multicolumn{1}{c}{} & \multicolumn{1}{c}{Rank-1} & \multicolumn{1}{c}{mAP} & \multicolumn{1}{c}{Rank-1} & \multicolumn{1}{c}{mAP} & \multicolumn{1}{c}{Rank-1} & \multicolumn{1}{c}{mAP} & \multicolumn{1}{c}{Rank-1} & \multicolumn{1}{c}{mAP} & \multicolumn{1}{c}{Rank-1} & \multicolumn{1}{c}{mAP} \bigstrut \\
    	\hline
    	\multicolumn{1}{l}{Baseline (ID cross-entropy only)} & \multicolumn{1}{c}{65.71} & \multicolumn{1}{c}{63.36} & \multicolumn{1}{c}{63.93} & \multicolumn{1}{c}{60.11} & \multicolumn{1}{c}{93.29 } & \multicolumn{1}{c}{83.06 } & \multicolumn{1}{c}{84.34} & \multicolumn{1}{c}{72.87} & \multicolumn{1}{c}{71.38 } & \multicolumn{1}{c}{47.08 } \bigstrut[t] \\
    	\multicolumn{1}{l}{Baseline + CPL} & \multicolumn{1}{c}{69.57} & \multicolumn{1}{c}{68.17} & \multicolumn{1}{c}{67.29} & \multicolumn{1}{c}{65.36} & \multicolumn{1}{c}{94.06} & \multicolumn{1}{c}{86.09} & \multicolumn{1}{c}{86.27} & \multicolumn{1}{c}{75.34}& \multicolumn{1}{c}{74.51} & \multicolumn{1}{c}{50.68} \bigstrut[b] \\
    	\hline

    	\multicolumn{1}{l}{Baseline + Center~\cite{wen2016discriminative}} & \multicolumn{1}{c}{64.57} & \multicolumn{1}{c}{63.03} &  \multicolumn{1}{c}{62.57} & \multicolumn{1}{c}{59.65} & \multicolumn{1}{c}{92.87} & \multicolumn{1}{c}{82.78} & \multicolumn{1}{c}{83.48} & \multicolumn{1}{c}{72.52}& \multicolumn{1}{c}{70.92} & \multicolumn{1}{c}{46.37} \bigstrut[t] \\
    	\multicolumn{1}{l}{Baseline + Center + CPL} & \multicolumn{1}{c}{67.36} & \multicolumn{1}{c}{66.28} & \multicolumn{1}{c}{66.43} & \multicolumn{1}{c}{63.96} & \multicolumn{1}{c}{93.41} & \multicolumn{1}{c}{84.70} & \multicolumn{1}{c}{85.14} & \multicolumn{1}{c}{74.89} & \multicolumn{1}{c}{72.51} & \multicolumn{1}{c}{49.27}\bigstrut[t] \\
    	\hline
    	
    	\multicolumn{1}{l}{Baseline + RLL~\cite{wang2019ranked}} & \multicolumn{1}{c}{67.86} & \multicolumn{1}{c}{66.71} & \multicolumn{1}{c}{63.79} & \multicolumn{1}{c}{62.84} & \multicolumn{1}{c}{93.79} & \multicolumn{1}{c}{84.65} & \multicolumn{1}{c}{86.18} & \multicolumn{1}{c}{75.25} & \multicolumn{1}{c}{72.45} & \multicolumn{1}{c}{48.80} \bigstrut[t] \\
    	\multicolumn{1}{l}{Baseline + RLL + CPL} & \multicolumn{1}{c}{71.29} & \multicolumn{1}{c}{\textbf{69.83}} & \multicolumn{1}{c}{\textbf{68.71}} & \multicolumn{1}{c}{\textbf{66.98}} & \multicolumn{1}{c}{94.09} & \multicolumn{1}{c}{86.21} & \multicolumn{1}{c}{86.58} & \multicolumn{1}{c}{76.05} & \multicolumn{1}{c}{73.23} & \multicolumn{1}{c}{50.25}\bigstrut[t] \\
    	\hline
    	
    	\multicolumn{1}{l}{Baseline + Triplet~\cite{schroff2015facenet}} & \multicolumn{1}{c}{68.93} & \multicolumn{1}{c}{66.82} & \multicolumn{1}{c}{65.21} & \multicolumn{1}{c}{62.90} & \multicolumn{1}{c}{93.68} & \multicolumn{1}{c}{84.90} & \multicolumn{1}{c}{85.19} & \multicolumn{1}{c}{75.07}& \multicolumn{1}{c}{73.80} & \multicolumn{1}{c}{50.07} \bigstrut[t] \\
    	\multicolumn{1}{l}{Baseline + Triplet + CPL} & \multicolumn{1}{c}{\textbf{71.86}} & \multicolumn{1}{c}{69.58} & \multicolumn{1}{c}{68.29} & \multicolumn{1}{c}{65.91} & \multicolumn{1}{c}{\textbf{94.66}} & \multicolumn{1}{c}{\textbf{86.96}} & \multicolumn{1}{c}{\textbf{87.34}} & \multicolumn{1}{c}{\textbf{76.41}}& \multicolumn{1}{c}{\textbf{75.00}} & \multicolumn{1}{c}{\textbf{51.84}} \bigstrut[t] \\
    	\hline
    	
    \end{tabular}
    }
    \caption{The CPL is complementary to other loss functions. The baseline is BoT with ID cross-entropy loss only. \textbf{Bold} fonts represent the best performance (in \%).}
	\label{tab:losses}
\end{center}
\end{table*}

\begin{table}[tb]
    \begin{center}
     \scalebox{0.78}{
    \begin{tabular}{c|cccc}
    \hline
    \cline{2-5}
    & Market1501 & Duke & CUHK03(L) & MSMT17 \\
    \hline\hline
    Random point. & $86.30$ & $76.39$ & $\textbf{69.71}$ & $51.15$ \\
    Farthest point    & $86.76$ & $76.18$ & $69.56$ & $51.57$ \\
    Sample mean & $86.59$ & $76.18$ & $69.33$ & $51.46$ \\
    Leave-one-out mean & $\textbf{86.96}$ & $\textbf{76.41}$ &$69.58$  & $\textbf{51.84}$ \\
    \hline
    \end{tabular}
    }
    \caption{The mAP (\%) comparison with different predictor target. \textbf{Bold} fonts represent the best performance.}
     \label{tab:target}
    \end{center}
\end{table}

\subsection{Ablation Study}
In this section, we perform an ablation study on several key factors.
Unless specified otherwise, the default setting in the experiment is BoT~\cite{luo2019bag} trained with ID cross-entropy loss, triplet loss, and our proposed CPL. 

\paragraph{Target of predictor}
In the CPL, the MLP predicts the center of the remaining same-class samples. However, theoretically, it is possible to use other prediction targets. In this ablation study, we conduct experiments to evaluate several alternatives. They are: (1) a randomly chosen same-class point (Random point), (2) a sample which is farthest to the current point (Farthest point), and (3) average of all the same-class samples (Sample mean). We compare them with the target used in the CPL, i.e., an average of remaining same-class samples (Leave-one-out mean). 

The results are shown in Table~\ref{tab:target}. We can see that the performance of these four different ways is comparable, and using the mean of remaining same-class samples is slightly better than others, so we use it as the target in CPL by default.




\paragraph{Complementarity of the CPL and other losses}
As mentioned in the introduction, the CPL is an intra-class loss that encourages certain distribution patterns of the same-class samples. It usually works with other loss functions. In this section, we conduct experiments to combine the CPL with other loss functions. The purpose of this experiment to check if the CPL can be beneficial in addition to the existing loss functions. 

The results are shown in Table~\ref{tab:losses}. As seen, the incorporation of the CPL leads to consistent performance improvement in almost all cases. This clearly shows that the CPL is complementary to the other loss functions.

\begin{table}[tb]
    \begin{center}
    \scalebox{0.78}{
    \begin{tabular}{c|c|cc|c}
    \hline
    Projection &
      \begin{tabular}[c]{@{}c@{}}Target BN\\ (output)\end{tabular} &
      \begin{tabular}[c]{@{}c@{}}Predictor BN\\ (hidden)\end{tabular} &
      \begin{tabular}[c]{@{}c@{}}Predictor BN\\ (output)\end{tabular} &
      \begin{tabular}[c]{@{}c@{}}mAP (\%)\end{tabular} \\ \hline \hline
    No Pred. & $-$ & $-$ & $-$ & $85.59$ \\ \hline  
    $2$-layer Pred.             & $-$ & $-$ & $-$ & $85.40$ \\ \hline
    $2$-layer Pred.              & $\checkmark$ & $-$ & $-$ & $86.77$ \\ \hline
    $2$-layer Pred.              & $\checkmark$ & $\checkmark$ & $-$ & $\textbf{86.96}$ \\ \hline
    $4$-layer Pred.              & $\checkmark$ & $\checkmark$ & $-$ & $86.89$ \\ \hline
    $2$-layer Pred.            & $\checkmark$ & $\checkmark$ & $\checkmark$ & $86.59$ \\ \hline
    \end{tabular}
    }
     \caption{The influence of batch normalization in CPL on Market1501.}
     \label{tab:projection_BN}
    \end{center}
    \vspace{-5mm}
\end{table}

\paragraph{The influence of batch-normalization (BN) layers}
We find that using batch-normalization can make the training more stable and lead to improved performance. We compare the different ways of applying BN to CPL: (1) without applying BN to calculate the target (Target BN) (2) only applying BN for target calculation. (3) applying BN for both target calculation and Predictor $f(\cdot;\theta)$ (we test applying BN to different positions in the predictor). The results are shown in Table~\ref{tab:projection_BN}. As seen, the use of BN is beneficial. However, the benefit mainly comes from applying BN to target calculation. Whether apply BN to the MLP predictor has little impact on the performance.


In addition, we also investigated the impact of the MLP structure. We test the performance by using a different number of layers in MLP. We find that the number of MLP layers in the predictor has little impact on the performance, and two layers are enough. Therefore, we choose a predictor with two layers, with its hidden node dimension as $512$.

\section{Conclusion}
In this work, we propose a new loss called the center prediction loss to strike the balance between intra-class variation reduction and flexible intra-class distribution modeling. The loss encourages a sample to be positioned in a location such that from it we can roughly predict the center of the same-class samples. 
We show that the proposed CPL allows greater freedom in data distribution families and it can also preserve the discrimination of learned embedding by avoiding positioning samples around the class boundaries. 
We carried out extensive experiments on various real-world ReID datasets to demonstrate that the proposed loss is a powerful alternative to many existing metric learning losses. Also, it can be complementary to these losses and achieve higher performance.



\small
\bibliographystyle{named}
\bibliography{ijcai21}

\end{document}